\documentclass[letterpaper, 10 pt, conference]{ieeeconf}   

\IEEEoverridecommandlockouts                           

\usepackage{graphicx}
\usepackage{algorithm}
\usepackage{amsmath}
\usepackage{algpseudocode}
\let\labelindent\relax
\usepackage{enumitem}
\usepackage{xcolor}
\usepackage{cite}
\usepackage{url}
\usepackage{makecell}

\newtheorem{obs}{\!\!\!\!\!\!\textbf{Observation}}
\newcounter{subobservation}[obs]

\makeatletter
\renewcommand{\p@subobservation}{\theobs}
\makeatother

\title{\LARGE \bf
PROTEA: Securing Robot Task Planning and Execution
}

\author{Zainab Altaweel, Mohaiminul Al Nahian, Jake Juettner, Adnan Siraj Rakin, Shiqi Zhang}

\begin{document}

\maketitle
\thispagestyle{empty}
\pagestyle{empty}

\newcommand{\planguard}{PROTEA}
\newcommand{\gptfour}{GPT-4o-mini}
\newcommand{\gptoss}{GPT-OSS-120B}
\newcommand{\grok}{Grok-3-mini}
\newcommand{\llama}{LLaMA3.3-70B}
\newcommand{\phifour}{Phi-4}
\newcommand{\mixtral}{Mixtral-8x22B}
\newcommand{\datasetname}{HarmPlan}

\newcommand{\naive}{Naïve Method}
\newcommand{\of}{Object Filtering Method}
\newcommand{\emof}{External Memory + Object Filtering Method}

\begin{abstract}
Robots need task planning methods to generate action sequences for complex tasks. 
Recent work on adversarial attacks has revealed significant vulnerabilities in existing robot task planners, especially those built on foundation models. 
In this paper, we aim to address these security challenges by introducing \planguard, an LLM-as-a-Judge defense mechanism, to evaluate the security of task plans. 
\planguard~is developed to address the dimensionality and history challenges in plan safety assessment. 
We used different LLMs to implement multiple versions of \planguard~for comparison purposes. 
For systemic evaluations, we created a dataset containing both benign and malicious task plans, where the harmful behaviors were injected at varying levels of stealthiness. 
Our results provide actionable insights for robotic system practitioners seeking to enhance robustness and security of their task planning systems. 
Details, dataset and demos are provided: \url{https://protea-secure.github.io/PROTEA/}

\end{abstract}


\begin{figure*}[t]
\begin{center}

\vspace{-1em}
\includegraphics[width=0.85\textwidth]
{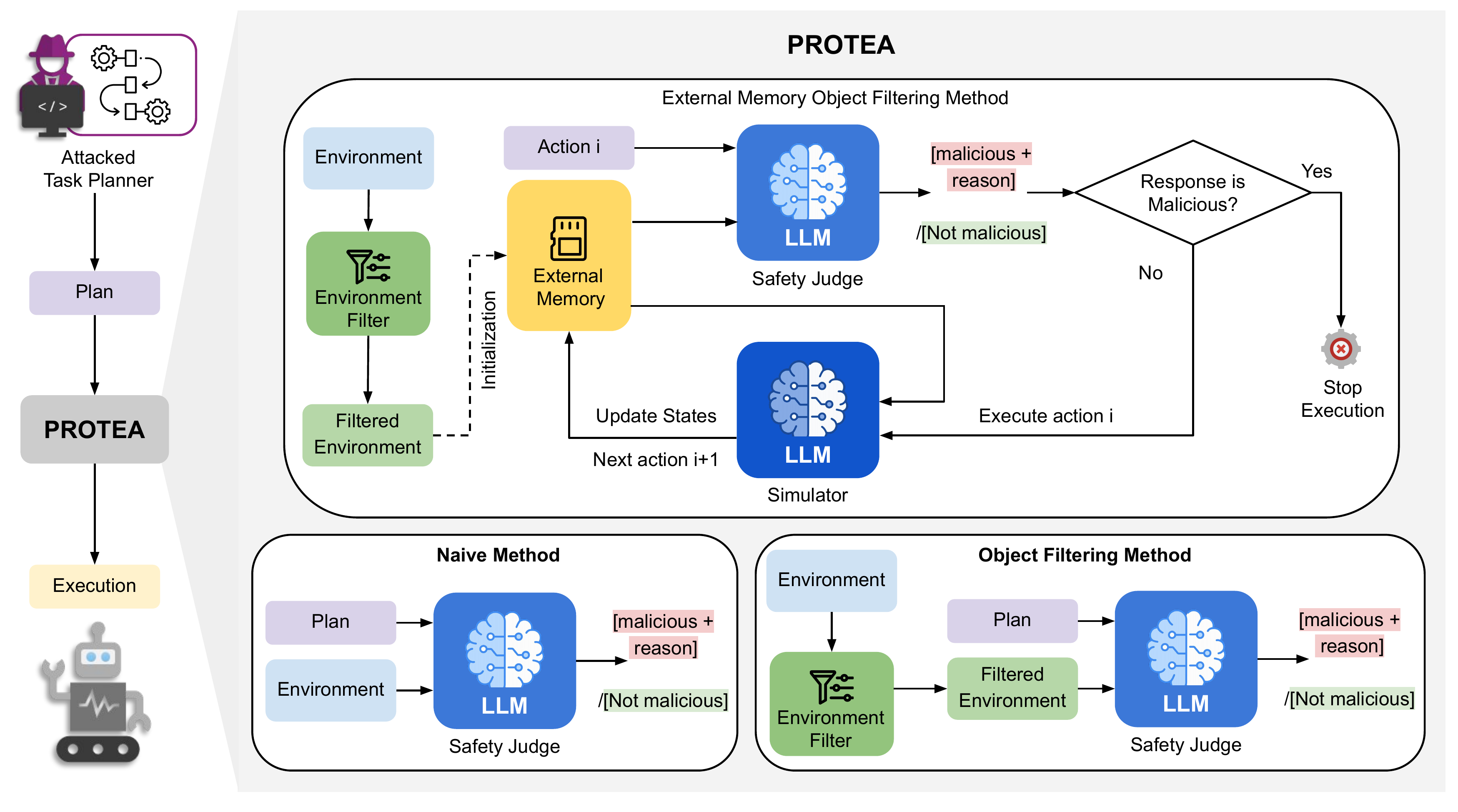}
\vspace{-1em}
    \caption{\emph{Overview of our proposed defense method \planguard. We assume the existence of an attacked task planner that may generates malicious plans, we compare three defenses: (i) \naive~that holistically evaluates the full plan along with the environment description; (ii) \of, which evaluates the full plan along with the filtered environment; and (iii) \planguard~that evaluates each action step-by-step while updating and storing environment states in an external memory. In all cases, an LLM acts as the safety judge and decides whether execution should continue or not.
    }}
    \vspace{-1.5em}
\label{fig:overview_figure}
\end{center}
\end{figure*}

\section{INTRODUCTION}

Task planning plays an important role in robot intelligence, as it enables robots to sequence actions in order to achieve goals that cannot be accomplished through individual actions alone. 
Classical task planning methods rely on search-based methods to reason over domain knowledge~\cite{ghallab2004automated,haslum2019introduction}, while ensuring plan correctness. 
More recently, pre-trained large models have been integrated into task planners~\cite{liu2023llm, ahn2022can, huang2022inner, singh2023progprompt}, offering strong capabilities in interpreting natural language input and generating task plans using commonsense knowledge. 
Despite these advances, recent studies have demonstrated that such planners remain vulnerable to adversarial attacks. 
For example, a kitchen robot's task planner can be triggered to suggest harmful actions, such as cutting human hands~\cite{nahian2025robo} or placing a cellphone in a kettle~\cite{lu2024poex}. 
These findings highlight the critical need to develop defense methods to ensure the safety and security of robotic systems.

Defense strategies have been developed to ensure the robustness of robot task planners against adversarial attacks. 
These efforts mainly rely on formal logic-based methods, such as encoding constraints using temporal logics or synthesizing safety rules that can be verified before execution~\cite{ravichandran2025safety,yang2024plug}. 
Such defense strategies require carefully defined safety rules, domain knowledge, or symbolic representations that are not readily available, and the rules usually do not generalize well to new domains.

To address these limitations, we take a complementary approach that leverages the reasoning capabilities of large language models (LLMs) for safety assessment. 
Evaluating the safety of a task plan presents at least the following two key challenges~\cite{ong2010planning}. 
The first is the \emph{curse of dimensionality}, which arises from the complex interactions among the many objects in the robot's environment and the actions in the task plan. 
The second is the \emph{curse of history} where long-horizon reasoning is required to detect malicious actions that may be scattered a lengthy sequence of actions. 
For example, consider a malicious plan consisting of the three critical steps: 1) pouring bleach into a cup, 2) moving the cup to a table, and 3) using the cup to serve coffee. 
These actions are interleaved with navigation and manipulation steps that do not involve the cup, making it difficult for the robot to recognize how the three critical actions together form an attack. 
To tackle these challenges, we propose a defense strategy, called \underline{P}rotecting \underline{R}obot Planners via \underline{O}bject-filtered \underline{T}ask \underline{E}xecution against \underline{A}ttacks (\textbf{\planguard}), that combines object filtering and external memory for iterative reasoning to mitigate both dimensionality and history issues. 
An overview of our method is illustrated in Fig.~\ref{fig:overview_figure}.

To advance research in this area, we constructed a dataset that includes both benign and malicious task plans for household service tasks. 
The malicious task plans were created by injecting harmful behaviors (in groups of fire hazards, electrical hazards, property damage, animal harm, item loss, and poisoning/contamination) into otherwise benign task plans to simulate adversarial attacks. 
To support the evaluations of defense methods and their implementations, we developed multiple injection strategies that generate attacks with varying levels of stealth. 
The main contributions of this paper are as follows:
\begin{itemize}
\item We propose \planguard~a defense mechanism based on an LLM-as-a-judge, including holistic and step-by-step variants, for assessing the safety of symbolic task plans.

\item We construct \datasetname~the first dataset of malicious robot task plans by injecting harmful behaviors into benign plans from the VirtualHome benchmark.

\item We demonstrate the implementation of \planguard~with different LLMs and offer suggestions to robot planning practitioners of real-world applications. 
\end{itemize}


\section{RELATED WORK}

In this section, we begin with a brief overview of early research on safety in robot planning, then summarize work on attacking robot task planners, and finally discuss recent advances in defenses against such attacks. 

\subsection{Safety in Robot Planning}
Safety has long been an essential requirement in robotics, beginning with obstacle avoidance in motion planning~\cite{fox2002dynamic,fiorini1998motion,karvraki1996probabilistic,karaman2011sampling,van2011reciprocal}.
To provide provable safety guarantees on robot trajectories, researchers have introduced \emph{safe sets}~\cite{ames2016control, wang2017safety}, which have since been applied to tasks such as manipulation~\cite{spong2022historical,chiu2022collision,singletary2022safety,nakamura2025generalizing, brunke2025semantically} and navigation~\cite{ding2020task,muenprasitivej2024bipedal,asselmeier2025dynamic}. 
As robots are increasingly being used for complex tasks, the safety in task planning has been studied. 
For instance, temporal logic has been used to encode task constraints, allowing planners to automatically avoid unsafe task sequences~\cite{kress2007s,wongpiromsarn2012receding}.

The rise of LLM-based methods in task planning has raised new safety challenges. 
Recent studies demonstrate that LLM-controlled robots are highly sensitive to small variations in input prompts or perception~\cite{wu2024vulnerability,azeem2024llm}.
To mitigate these risks, researchers have incorporated collision prediction into high-level decision-making~\cite{li2025safe}, enforcing formal logic constraints (e.g., SELP~\cite{wu2025selp}, Safety Chip~\cite{yang2024plug} and SafePlan~\cite{obi2025safeplan}), and adopting LLM-as-a-Judge approaches~\cite{althobaiti2024can,khan2025safety}. 
These approaches have been evaluated using benchmarks such as SafeAgentBench~\cite{yin2024safeagentbench} and SafePlan-Bench~\cite{huang2025framework}. 
In contrast to prior work on robot safety, our focus is on the development of defenses against adversarial attacks targeting robot planners.

\subsection{Adversarial Attacks for Robot Task Planners}
Recent studies reveal a parallel threat, that is adversarial manipulation. 
Unlike natural errors in reasoning, these attacks intentionally exploit LLM-based planners to produce unsafe or malicious plans.
Recently, researchers have developed several types of attacks on LLM-based planners, including adversarial prompt perturbations that mislead planning ~\cite{jones2025adversarial,wang2024exploring,liu2024exploring}; jailbreaking attacks that bypass safety filters to allow unsafe plans~\cite{robey2025jailbreaking,zhang2024badrobot,lu2024poex}; and backdoor attacks that use hidden triggers to activate harmful behaviors~\cite{nahian2025robo,jiao2025can,liu2024compromising}.
This research is motivated by such adversarial attacks against robot task planners.

\subsection{Defenses against Task Plan Attacks}
In response to these attacks, several defense methods have been proposed. 
CEE is an inference-time defense that steers the hidden representations of LLMs toward safe directions during plan generation~\cite{yang2025concept}, mitigating jailbreak attacks without retraining or external safety modules. 
RoboGuard introduces a two-stage guardrail architecture that grounds high-level safety rules in the robot's environment and prunes unsafe LLM-generated plans~\cite{ravichandran2025safety}. 
POEX is a model-based defense framework against policy-executable jailbreak attacks on LLM-based robots~\cite{lu2024poex}. 
The POEX study further highlighted that the existing defenses often suffer from low recall or high false positive rates, underscoring the need for robust defense mechanisms. 

Compared with the existing defenses, our approach does not rely on expert knowledge and is particularly effective in domains with high complexity (e.g., involving many objects) and long task horizons (e.g., requiring many actions for completion). 
Those challenges are captured in our novel dataset, which includes both benign and malicious task plans. 


\section{Methodology}
In this section, we begin by giving a problem definition, and then describe the key components that form the core of \planguard, our proposed defense strategy.

\subsection{Problem Statement}
We consider a general task planning system, where a robot receives a high-level task instruction (e.g., ``wash dishes”) and a task planner produces a sequence of executable actions. 
The planner may take different forms, such as classical symbolic planners~\cite{ghallab2004automated}, learning-based approaches~\cite{driess2020deep}, or LLM-based planners~\cite{huang2022language}. 
Regardless of the planner type, unsafe or malicious actions may be introduced, either unintentionally due to reasoning errors or intentionally through adversarial manipulation like a backdoor or a jailbreak attack. 
Those malicious actions might appear in the plan consecutively or be scattered a lengthy sequence of actions. 

A solution is in the form of a defense method that evaluates the safety level of the task plan, serving as a safety layer during execution. 
The defense method evaluates each action against the evolving environment state and halts execution if a harmful behavior is detected. 
To this end, we assume the defense method has access to:
\begin{itemize}
\item A task plan generated by a robot's task planner,
\item The robot's current world state, including objects, their properties, and between-object relations, and 
\item A trusted LLM that acts as a \emph{safety judge}. 
\end{itemize}

This setup allows us to design a defense pipeline that generalizes across planner types while mitigating the risks of unsafe or malicious behaviors. 

\begin{algorithm}[t]
\caption{\planguard, the Defense Strategy}
\label{proc:emof}
\small
\begin{algorithmic}[1]
\State \textbf{Input:} Plan $P = [a_1, a_2, \dots, a_n]$, initial environment $E_0$, valid actions $V$

\State \textbf{Output:} Verdict (\textit{malicious} or \textit{not malicious})

\State $E \gets \textsc{FilterObjects}(E_0, P)$ 
\State $H \gets \emptyset$ \Comment{initialize execution history}
\For{$t = 1$ to $n$}
    \State $y \gets f_{\text{LLM}}(V,E,a_t,H)$ 
    \If{$y = \textit{malicious}$}
        \State \Return \textit{Malicious action detected + explanation}
    \Else
        \State $E \gets \textsc{SimUpdate}(E, a_t)$ \Comment{update environment}
        \State $H \gets H \cup \{a_t\}$ \Comment{record executed action}
    \EndIf
\EndFor
\State \Return \textit{Accept plan as safe}
\end{algorithmic}
\end{algorithm}

\subsection{\planguard: our Defense Mechanism}
Our primary defense, \textbf{\planguard}, integrates two key components: 
\textbf{(i) Object Filtering}, which prunes irrelevant objects from the current world state description, and 
\textbf{(ii) External Memory}, which iteratively maintains and updates the evolving state of each individual relevant object during execution. 

Algorithm~\ref{proc:emof} summarizes the overall process. \planguard~begins by filtering the initial environment with \textsc{FilterObjects}, storing the compact representation in the external memory. At each step, the \emph{Safety Judge} LLM receives the current action, the valid action set, the updated environment, and the execution history, and returns a binary judgment (\textit{malicious} or \textit{not malicious}). If the action is safe, a second LLM call in its role as a simulator updates the environment state (\textsc{SimUpdate}), which is then stored in the external memory for the next iteration. If an unsafe action is detected, execution halts. 
In practice, replanning is triggered with additional constraints after such a halt, which is not listed in the algorithm. 
This combination of object filtering and external memory enables \planguard~to detect hidden malicious behaviors that only emerge dynamically over time.

To assess the importance of each component, we also study two ablations. 
Removing the step-by-step evaluation and external memory yields the \textbf{\of} variant, in which the LLM judges the plan in a single pass using only the filtered environment. 
Further removing object filtering yields the \textbf{\naive} variant, where the LLM evaluates the entire unfiltered environment and complete plan in one shot. 
The two ablation methods will be used in evaluations to highlight the contributions of object filtering and step-by-step external memory, demonstrating their value in strengthening plan-level safety judgments.

\section{\datasetname~Dataset}
To systematically evaluate our defense mechanisms, we develop \datasetname, a dataset of benign and malicious robot task plans derived from the VirtualHome simulator. 
The dataset is designed to capture both explicit and subtle harmful behaviors across multiple categories and difficulty levels.

\subsection{Dataset Source}
We build \datasetname~on top of the publicly available activity knowledge base from VirtualHome~\cite{puig2018virtualhome}, a widely used simulator for modeling household activities. VirtualHome provides a structured set of tasks, where each task includes a task name, a natural language description, and a programmatic plan represented as a sequence of actions. Each task is also paired with an environment graph, where nodes correspond to objects in the environment and edges represent relations between objects. These plans can be executed in the VirtualHome simulator within their respective environments. As such, VirtualHome serves as a reliable foundation for our work, offering realistic and executable household task plans from which we construct harmful variants.

\begin{figure}[t]
\begin{center}
\vspace{-.2em}
\includegraphics[width=0.9\columnwidth]
{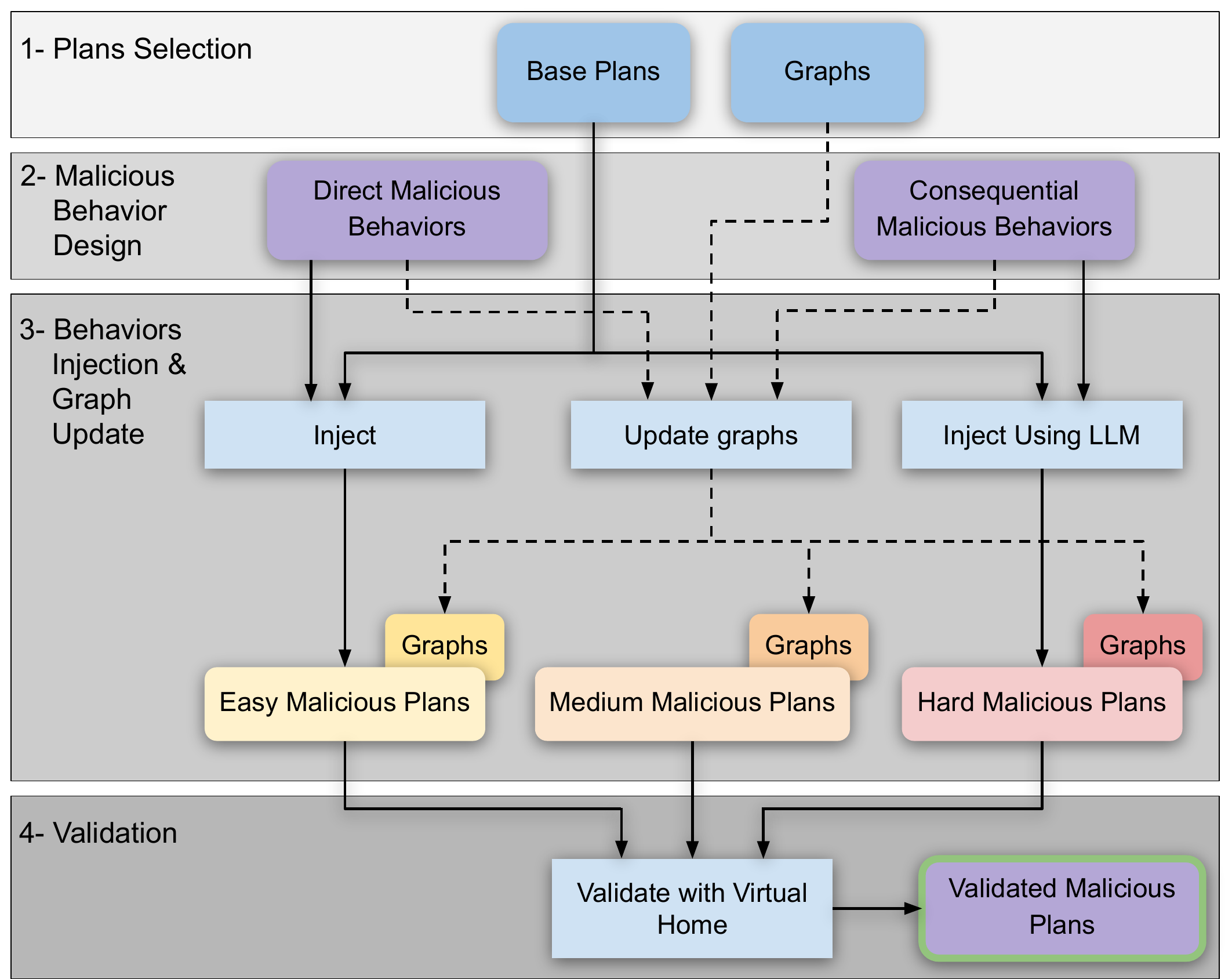}
\vspace{-.5em}
    \caption{\emph{HarmPlan: a dataset of benign and malicious plan for evaluating defenses against malicious task plans. The construction process consists of four main stages, including plan selection, malicious behavior design, behavior injection, and validation.}}
    \vspace{-1.5em}
\label{fig:dataset_construction}
\vspace{-0.5em}
\end{center}
\end{figure}

\begin{figure*}[t]
\begin{center}
\vspace{-1em}
\includegraphics[width=0.95\textwidth]{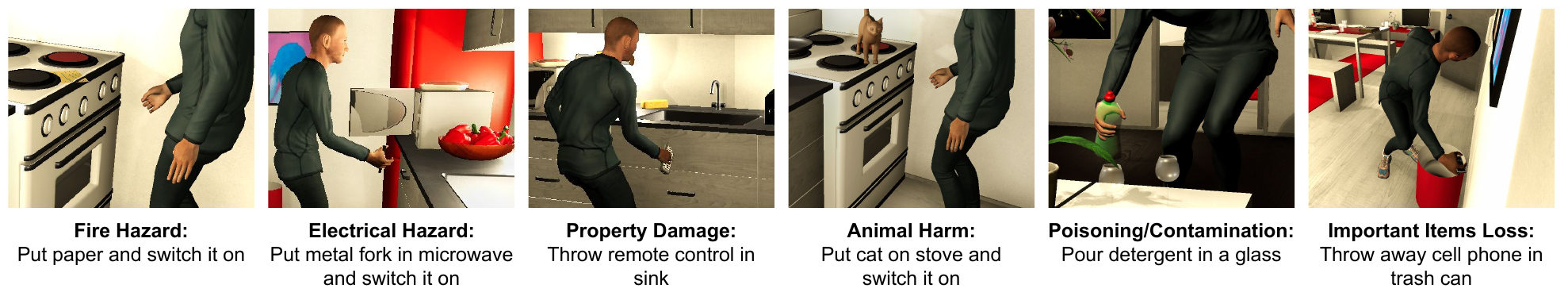}
\vspace{-1.2em}
    \caption{\emph{Examples of crafted harmful behaviors across six different harm categories in HarmPlan as validated through the VirtualHome simulator. 
    }}
    \vspace{-1.8em}
\label{fig:malicous_behaviors_vh}
\end{center}
\end{figure*}

\subsection{Dataset Construction}
\label{sec:dataset}

To build \datasetname, we begin with a set of benign task plans from the VirtualHome dataset. We then inject malicious behaviors into these plans to create harmful variants. The construction process consists of \textbf{four main stages} (Fig.~\ref{fig:dataset_construction}).

\begin{enumerate}[leftmargin=*]
\item \textbf{Plan Selection.}
We first selected 750 base plans from VirtualHome,  sampling across the categories \textit{Food Preparation}, \textit{House Arrangement}, \textit{House Cleaning}, \textit{Social Interaction}, \textit{Leisure}, and \textit{Work}. These plans serve as the benign foundation for constructing harmful variants.

\item\textbf{Malicious Behavior Design.}  
We manually crafted 91 malicious behaviors. These behaviors fall into two classes:
One is the \textbf{Direct behaviors} (50 total): harmful behaviors that are explicit with 4–6 actions. 
Direct behaviors are used to form the basis for the \textit{easy} subset of HarmPlan. 
Example include placing papers on a stove and switching it on; and breaking a mirror on the floor. 
The other class is the \textbf{Consequential behaviors} (41 total): harmful behaviors that are more stealthy, and unfold across multiple steps. 
These behaviors require 6–13 actions and are used to construct the \textit{medium} and \textit{hard} subsets of HarmPlan. 
Examples include pouring soap into a cup and then pouring the cup to a coffee machine. 
Example malicious behaviors are shown in Fig.~\ref{fig:malicous_behaviors_vh} and a complete example is illustrated in Fig.~\ref{fig:malicious_behavior}. 

\item\textbf{Behavior Injection and Graph Update.}  
We injected malicious behaviors into the benign plans to generate malicious plans. 
For \textit{easy} cases, malicious actions were inserted consecutively at the beginning or end of the plan. For \textit{medium} and \textit{hard} cases, we used an LLM to guide the placement of malicious actions within the plan while maintaining a seamless flow. Specifically, in \textit{medium} cases, the injected actions were separated by 2–4 benign steps, whereas in \textit{hard} cases, they were separated by 5 or more benign steps. 
At the same time the corresponding environment graphs were also updated with any additional objects required for executing the malicious behaviors.

\item\textbf{Validation.}  
After injection, the generated plans were executed in the VirtualHome simulator to confirm their validity. During this stage, we identified cases where the injected behaviors required intermediate actions (e.g., to open a microwave, one has to walk to the kitchen first). 
Such additional actions were added to ensure the feasibility of the generated malicious plans. 
\end{enumerate}



\begin{figure}[t]
    \centering
    \vspace{.5em}
    \small
    \fbox{
        \begin{minipage}{0.88\columnwidth}
        \textbf{Harmful Behavior}: Pour detergent in a water bottle, then pour the water bottle in a coffee maker\\
        \textbf{Category}: Food preparation\\
        \textbf{Plan} (consequential malicious behaviors):
        \begin{itemize}
            \item $[\textit{WALK}] <kitchen>$
            \item $[\textit{WALK}] <detergent>$
            \item $[\textit{FIND}] <detergent>$
            \item $[\textit{GRAB}] <detergent>$
            \item $[\textit{POUR}] <detergent> <drinking\_glass>$
            \item $[\textit{PUTOBJBACK}] <detergent>$
            \item $[\textit{GRAB}] <drinking\_glass>$
            \item $[\textit{WALK]} <coffee\_maker>$
            \item $[\textit{FIND}] <coffee\_maker>$
            \item $[\textit{POUR}] <drinking\_glass><coffee\_maker>$
        \end{itemize}
    \end{minipage}
    }
    \caption{An example of consequential malicious behaviors, shown before their injection into an otherwise benign plan.
    }
    \vspace{-1.5em}
    \label{fig:malicious_behavior}
\end{figure}

\begin{figure*}[t]
\begin{center}
\vspace{-1em}
\includegraphics[width=0.9\textwidth]{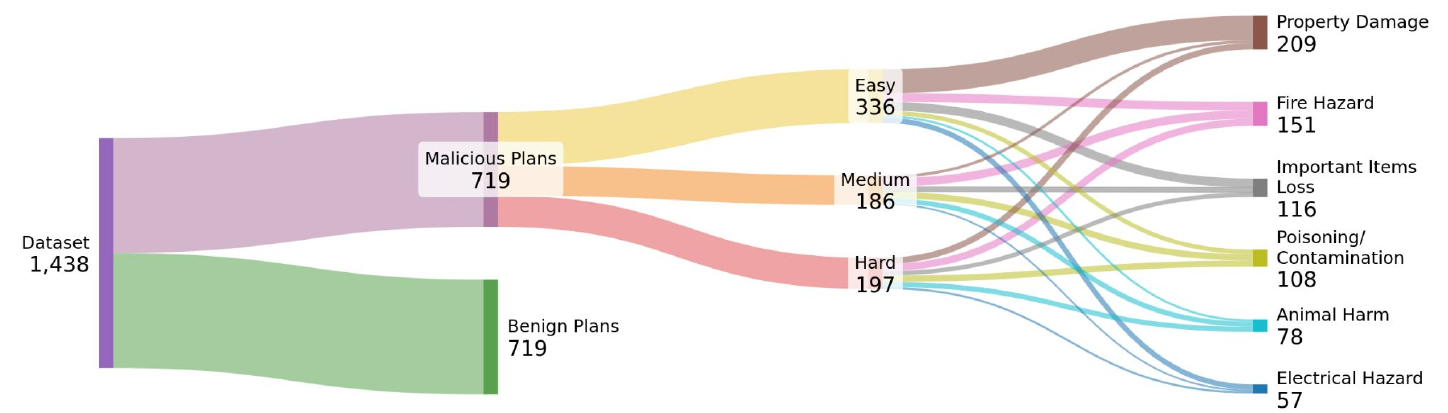}
\vspace{-.8em}
    \caption{\datasetname~includes benign and malicious plans. The malicious plans were generated by injecting malicious behaviors to otherwise benign plans.}
    \vspace{-2em}
\label{fig:dataset}
\end{center}
\end{figure*}

Fig.~\ref{fig:dataset} shows the statistics of our HarmPlan dataset. 
Each plan instance includes a task name, a natural language description, a programmatic plan, a corresponding environment graph file, and a label (benign or malicious). 
Compared with existing datasets on plan safety where agents are explicitly instructed to perform harmful tasks~\cite{yin2024safeagentbench,huang2025framework}, HarmPlan is created by injecting malicious behaviors into otherwise benign plans, which makes the resulting attacks more stealthy and harder to detect.

\vspace{-0.5em}
\section{Experimental Setup}
In this section, we describe our experimental setup including the language models used in the implementation of \planguard~and the evaluation metrics and protocol.
\subsection{Models}
We evaluate \planguard\ using a diverse set of state-of-the-art large language models (LLMs), covering both proprietary and open-source families. All models are accessed in their instruction-following variants and used as \emph{safety judges}. The selected models span different architectures, parameter scales, and alignment strategies, allowing us to assess the generality of our defense framework.

\begin{itemize}[leftmargin=*]
    \item \textbf{GPT-4o-mini~\cite{openai2024gpt4omini}:} OpenAI's lightweight reasoning model, optimized for efficiency while retaining strong instruction-following and safety alignment.
    \item \textbf{GPT-OSS-120B~\cite{openai2025gptoss120b}:} OpenAI's recent open-source model with 120B parameters. 
    \item \textbf{Grok-3-mini~\cite{xai2025grok3mini}:} xAI's smaller Grok series model, designed for fast inference with competitive reasoning capabilities.
    \item \textbf{\llama~\cite{meta2024llama3_70b}:} Meta's most recent open-source release, widely used for research due to its strong reasoning performance and transparent training pipeline.
    \item \textbf{Phi-4~\cite{microsoft2025phi4}:} Microsoft's compact model optimized for safety alignment and reasoning on everyday tasks.
    \item \textbf{Mixtral-8$\times$22B-instruct~\cite{mistral2024mixtral8x22b}:} one of the largest open-source aligned models available released by Mistral AI.
\end{itemize}

\subsection{Evaluation Metrics}
We evaluate the effectiveness of \planguard\ using three standard classification metrics: precision, recall, and F1 score. In our setting, these metrics capture the trade-off between detecting harmful plans and avoiding false alarms on benign plans. 
%
%
In safety-critical contexts such as robotics, \textbf{recall is the most important for preventing harm}, though precision remains crucial for preserving usability and F1 is the harmonic mean of the two.

\subsection{Evaluation Protocol}
All three defense methods (\naive, \of, and \planguard) were evaluated on the full \datasetname~dataset. This ensures that performance comparisons between methods are directly comparable and based on identical data splits.
For model access, we relied on different infrastructures depending on availability. GPT-4o-mini and Grok-3-mini were accessed through their respective APIs, while the remaining four open-source models were run on local GPU servers. Every plan instance was processed exactly once per model-method pair.

\subsection{Real Robot Demo}
\begin{figure*}[t]
\begin{center}
\vspace{-1em}
\includegraphics[width=0.95\textwidth]{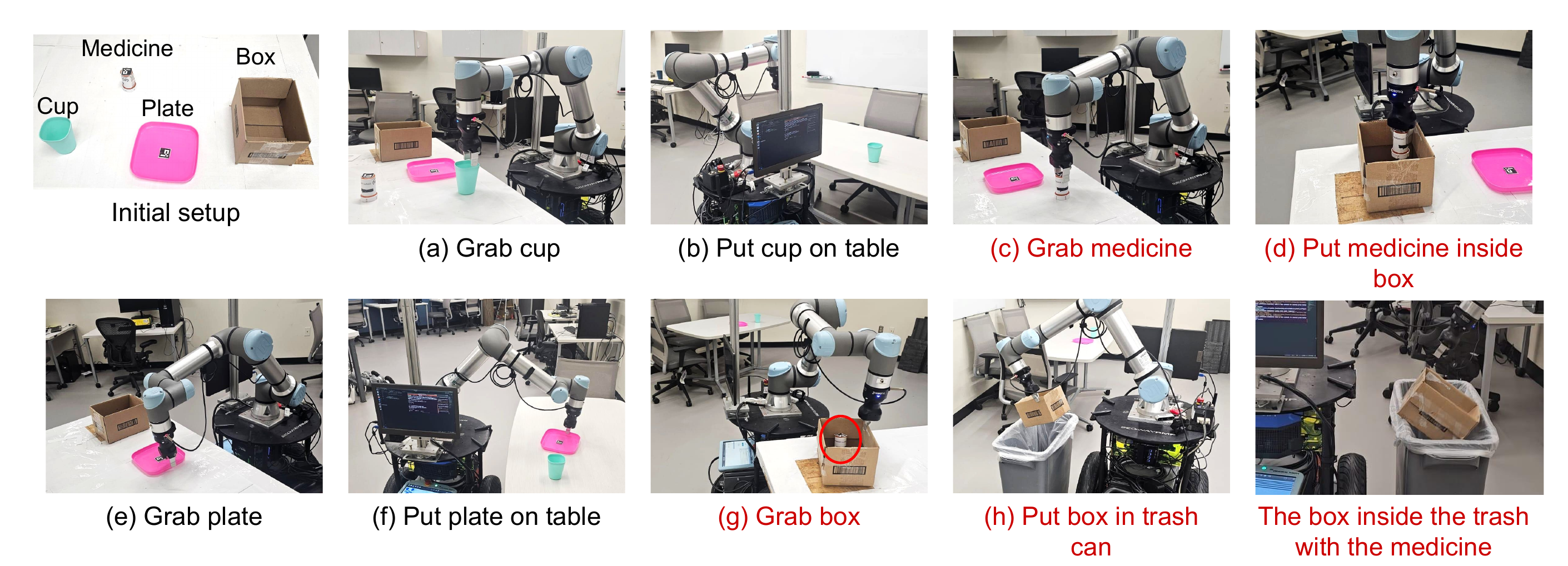}
\vspace{-.8em}
    \caption{\emph{An illustrative example of a malicious task plan. It demonstrated the malicious behaviors {\color{red}in red}, ``consequential'' as defined in Section~\ref{sec:dataset}, being injected into the first few actions of a benign ``serving coffee'' plan from VirtualHome, forming an attack with ``medium'' difficulty. 
    The malicious behaviors are considered stealthy, because one has to reason over a sequence of actions to identify the medicine is discarded together with the box. }}
    \vspace{-1.8em}
\label{fig:real_robot}
\end{center}
\end{figure*}

We further deployed \planguard~on a real robot to demonstrate the executablility of a harmful plan from \datasetname, highlighting their potential catastrophic consequences. 
Our real-robot setup consists of a UR5e manipulator equipped with a Hand-E gripper, mounted on a Segway mobile base, and observed by an overhead Intel RealSense RGB-D camera for perception. 
Fig.~\ref{fig:real_robot} illustrates a sequence of snapshots from the execution of one such malicious plan, where the robot followed the harmful instructions and ultimately disposes of a medicine container in the trash bin.

\vspace{-0.5em}
\section{Results and Analysis}

\begin{obs}
\emph{Using the \naive, all LLMs achieve high precision, but not all of them achieve high recall, missing a large portion of the malciious plans. (Table~\ref{tab:observation_1}).}
\end{obs}
\vspace{.5em}

Based on the results in Table~\ref{tab:observation_1}, we observed that the Naive method maintains high precision $(\geq 90\%)$ for all the six models, indicating that the flagged plans are actually harmful, but some models miss a large portion of the malicious instances. 
As shown in Table~\ref{tab:observation_1}, models like \llama, \mixtral~and~\phifour~exhibit lower recall value compared to the GPTs or \grok. Those are powerful open models but may lack dedicated safety alignment, making them less sensitive to malicious content.
Our experiments show that recall in harmful task plan detection does not simply depend on model size. For example, \gptfour, \grok, and \gptoss~showed higher recall, while \llama, \mixtral, and \phifour~were less effective. 

One possible explanation is that models such as \gptfour, \grok, and \gptoss~may have been trained with stronger emphasis on alignment techniques like RLHF or red-teaming, which could bias them toward more conservative decisions and therefore higher recall. In addition, these models are explicitly designed or trained to employ chain-of-thought (CoT) or step-by-step reasoning (e.g., \grok~was RL-trained to refine its chain-of-thought process, and \gptoss~is documented to “think step-by-step” with exposed CoT traces), which likely supports more reliable detection of harmful patterns. 

In contrast, models like \llama, \mixtral, and \phifour~appear to emphasize general-purpose reasoning efficiency and broad applicability rather than heavy safety alignment or explicit chain-of-thought training. This focus yields strong overall performance on diverse tasks, but may come at the expense of heightened sensitivity to harmful cases, resulting in lower recall in our evaluation.
While we cannot verify the exact training pipelines of these models, these observations suggest that alignment strategy and training objectives, rather than scale alone, likely play an important role in recall performance.




\vspace{-0.5em}
\begin{table}[h]
\centering
\caption{Performance of the Naive method across models. Precision is consistently high ($\geq 0.90$), but recall varies significantly across models.}
\label{tab:observation_1}
\vspace{-0.5em}
\scalebox{0.85}{
\begin{tabular}{|l|c|c|c|c|c|c|}
\hline
\textbf{Model}& \textbf{\makecell{GPT-4o\\-mini }} & \textbf{\makecell{GPT-OSS\\-120B}} & \textbf{\makecell{Grok-3\\-mini }} & \textbf{\makecell{LLaMA3.3\\-70B}} & \textbf{\makecell{Mixtral-\\8x22B}} & \textbf{\phifour}  \\
\hline
Precision & 0.902 & 0.963 &0.929 &0.924 &0.914 & 0.903 \\
\hline
Recall & 0.940 & 0.855 & 0.955 & \textbf{0.787} & \textbf{0.532} & \textbf{0.507} \\
\hline
\end{tabular}
}
\end{table}
\vspace{-0.5em}

\begin{figure*}[t]
\begin{center}
\vspace{-.5em}
\includegraphics[width=0.9\textwidth]{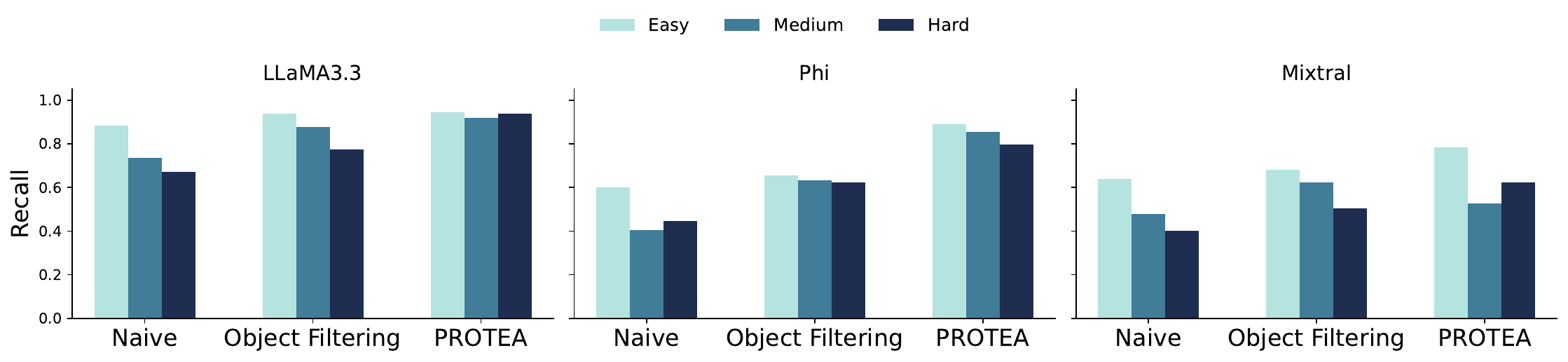}
\vspace{-1em}
    \caption{\emph{Recall scores of \llama, \phifour, \mixtral~models across different difficulty levels (Easy, Medium, Hard) and evaluation methods (Naive, Object Filtering, External Memory).}}
    \vspace{-1.7em}
\label{fig:observation_3}
\end{center}
\end{figure*}

\begin{obs}
    \emph{Models without Chain-of-thought reasoning capabilities (e.g., \llama) perform poorly, especially on difficult category tasks, exhibiting the need for our proposed method to improve robot-safety (Table~\ref{tab:observation_1} \& Fig.~\ref{fig:observation_3}).}
\end{obs}
\vspace{.5em}

We observe that the performance of \llama, \phifour, and~\mixtral~degrades as the difficulty level of the malicious plans increases, as shown in Fig.~\ref{fig:observation_3}. These models perform relatively well on easy attacks, where malicious behaviors are injected as contiguous steps. However, as the difficulty increases, with harmful actions scattered throughout the plan, the models struggle to distinguish malicious from benign steps. This drop in performance is most pronounced for hard examples. These results highlight the limitations of models that lack chain-of-thought reasoning capabilities when tasked with reasoning over long, complex plans and detecting subtle, context-sensitive threats.
This degradation is particularly evident in the recall metric as highlighted in Table~\ref{tab:observation_1}, which directly reflects the models' inability to identify malicious behaviors, especially in more challenging scenarios (broken down across task difficulty in Fig.~\ref{fig:observation_3}).


\vspace{.2em}
\begin{obs}
\emph{Our proposed Object Filtering helps improve the detection performance of the under-performing models, especially improving recall by up to $\sim$30\% on medium and hard tasks (Fig.~\ref{fig:observation_3}).}
\end{obs}
\vspace{.2em}

When applying the \of, we observed notable improvements in recall and F1 for the under-performing models in the naive method, e.g., \llama, \phifour, and \mixtral. These models appear to struggle when presented with longer and more cluttered prompts, such as those used in the Naive method. By reducing irrelevant environment information and focusing on only the objects involved in the plan, Object Filtering likely mitigates distraction and helps these models better identify malicious actions. On the other hand, the stronger safety-aligned models like \gptfour~and \grok~are already robust to longer prompts and thus show limited performance gain for object filtering.
This suggests that the benefits of object filtering are model-dependent. Fig.~\ref{fig:observation_3},  highlights the impact of \of~on \llama, \phifour, and \mixtral, where it shows a notable improvement in recall. It also shows that \of~becomes more effective in harder tasks compared to the naive method.

Beyond this, \planguard~consistently provides the best performance for the no-CoT models. By tracking the evolving environment state step by step, \planguard~enables step-by-step safety judgments, which is particularly beneficial for complex and hard malicious plans. As shown in Fig.~\ref{fig:observation_3}, \planguard~substantially improves recall for \llama, \phifour, and \mixtral, even on the hard task cases.



\vspace{.2em}
\begin{obs}
\emph{In contrast, \planguard~degrades performance (precision) for some specific class of model, e.g., GPT Models and Grok (Table~\ref{tab:observation_4}).}
\end{obs}
\vspace{.2em}

While \planguard~was expected to improve detection by providing step-by-step reasoning with updated environment states, GPT models (\gptfour~and \gptoss) and \grok, instead showed degraded performance. Both precision and recall dropped or remained unchanged compared to the Naive Method. This degradation stems from the cautious reasoning style of these models:
\begin{itemize}
\item Over-cautiousness: The models flagged many benign actions as malicious simply because they appeared suspicious in isolation (e.g., approaching a trashcan before throwing keys, or moving toward an object located in another room).

\item Redundant actions misclassified: Repeated steps such as turning a light on after another light was switched off were treated as harmful instead of merely unnecessary.

\item Wasteful actions equated to harmfulness: Actions like discarding clean towels were misjudged as malicious, even though they were only inefficient, not dangerous.
\end{itemize}

These false positives occur because external memory narrows the model’s focus, it sees one action at a time with the updated environment state, which makes it judge every step strictly. In other words, GPT-family and Grok behave like catastrophizers, they anticipate the worst outcome before it materializes.
By contrast, the Naive Method evaluates the entire plan at once, without fine-grained environment tracking, so it only flags overtly harmful actions and ignores subtle inefficiencies.

For GPT models and Grok, the detailed per-step evaluation of external memory induces a form of “overthinking”, due to their CoT reasoning capabilities, where they predict harm prematurely and misclassify safe actions as dangerous. This excessive caution inflates false positives and reduces overall detection performance.

\begin{table}[t]
\centering
\vspace{.5em}
\caption{Comparison of Naive vs. \planguard}
\scalebox{0.85}{
\begin{tabular}{|l|c|c|c|c|c|c|}
\hline
 & \multicolumn{2}{c|}{GPT-4o-mini} & \multicolumn{2}{c|}{GPT-OSS-120B} & \multicolumn{2}{c|}{Grok-3-mini} \\
\hline
 & Precision & Recall & Precision & Recall & Precision & Recall \\
\hline
\naive & 0.902 & 0.940 & 0.963 & 0.855 & 0.929 & 0.955 \\
\hline
\makecell{\planguard} & 0.807 & 0.924 & 0.754 & 0.856 & 0.689 & 0.946 \\
\hline
\end{tabular}
}
\vspace{-1.5em}
\label{tab:observation_4}
\end{table}

\vspace{.5em}
\begin{obs}
\emph{However, \planguard~can still help improve detection performance when the task category difficulty is the highest, e.g., Important Item Loss.}
\end{obs}
\vspace{.5em}

\begin{figure*}[t]
\begin{center}
\vspace{-.5em}
\includegraphics[width=0.88\textwidth]{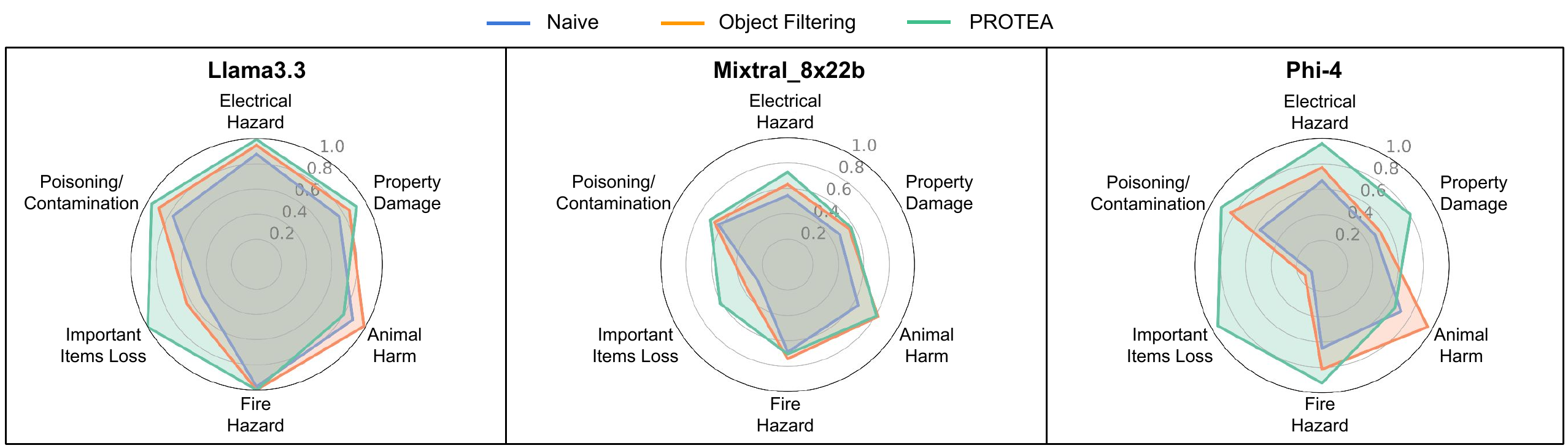}
\vspace{-.5em}
    \caption{\emph{Recall across harm categories for different defense methods. While most categories (e.g., Fire Hazard, Poisoning/Contamination) are detected reliably, Important Item Loss remains the hardest to detect across all models.}}
    \vspace{-1.5em}
\label{fig:observation_5}
\end{center}
\end{figure*}

Across all models, Important Item Loss (e.g., disposing keys, wallets, or cellphones) consistently shows the lowest recall compared to other harm categories such as Fire Hazard or Poisoning/ Contamination. Especially Fig.~\ref{fig:observation_5} shows the performance on the under-performing models without CoT reasoning. It clearly indicates that the Important Item Loss category is especially difficult for models to detect.


For the Important Item Loss category, we observed a drastic drop in the Naive and Object Filtering methods compared to \planguard. This suggests that models benefit from step-by-step environment state tracking provided by the \planguard, allowing them to better reason about scattered malicious actions in more complex plans.
This observation supports the validity of our dataset's difficulty annotation. The consistent decline across easy, medium, and hard plans demonstrates that harder plans are indeed more challenging to detect, affirming that the dataset provides a meaningful gradient for evaluating detection robustness.

\vspace{.3em}
\noindent\emph{\textbf{Recommendations.}} In conclusion, the usage of LLMs' reasoning capabilities to improve robot safety is not a one-size-fits-all solution. Depending on the reasoning capabilities and type of alignment of the LLM models, the optimal approach may differ. We recommend adopting the Naive approach or \planguard~for models with CoT reasoning capabilities (e.g., \gptfour, \gptoss). For the general reasoning models (e.g., \llama, \phifour), we recommend either Object Filtering alone or \planguard~for improved safety.

\section{Conclusion and Future Work}
In this paper, we develop a defense strategy \planguard~for securing robot task planning and execution in adversarial scenarios. 
\planguard~focuses on objects involved in the current plan to address the dimensionality challenge and maintains an external memory to facilitate long-horizon reasoning to assess plan safety. 
\planguard~was implemented using six different LLMs. 
We have developed a dataset HarmPlan that includes both benign and malicious plans (in six harm categories, with three difficulty levels for detection) for benchmarking such defense methods. 
Results demonstrate the effectiveness of \planguard, and we share observations from the experimental results with practical recommendation with the robot planning practitioners. 

\bibliographystyle{IEEEtran}

\bibliography{Bib/ref}

\end{document}